# A Data Quarantine Model to Secure Data in Edge Computing

**Poornima Mahadevappa[1], Raja Kumar Murugesan[2]**
[1,2]School of Computer Science and Engineering, Taylor's University, Malaysia

| Article Info | ABSTRACT |
|---|---|
| *Article history:*<br><br>Received May 19, 2021<br>Revised Jun 22, 2021<br>Accepted Aug 15, 2021 | Edge computing provides an agile data processing platform for latency-sensitive and communication-intensive applications through a decentralized cloud and geographically distributed edge nodes. Gaining centralized control over the edge nodes can be challenging due to security issues and threats. Among several security issues, data integrity attacks can lead to inconsistent data and intrude edge data analytics. Further intensification of the attack makes it challenging to mitigate and identify the root cause. Therefore, this paper proposes a new concept of data quarantine model to mitigate data integrity attacks by quarantining intruders. The efficient security solutions in cloud, ad-hoc networks, and computer systems using quarantine have motivated adopting it in edge computing. The data acquisition edge nodes identify the intruders and quarantine all the suspected devices through dimensionality reduction. During quarantine, the proposed concept builds the reputation scores to determine the falsely identified legitimate devices and sanitize their affected data to regain data integrity. As a preliminary investigation, this work identifies an appropriate machine learning method, Linear Discriminant Analysis (LDA), for dimensionality reduction. The LDA results in 72.83% quarantine accuracy and 0.9 seconds training time, which is efficient than other state-of-the-art methods. In future, this would be implemented and validated with ground truth data. |
| *Keywords:*<br><br>Edge computing<br>Data security<br>Data integrity<br>Quarantine<br>Data analysis | |
| |  |

*Corresponding Author:*

Raja Kumar Murugesan,
School of Computer Science and Engineering,
Taylor's University,
1, Jalan Taylors, 47500 Subang Jaya, Selangor, Malaysia.
Email: rajakumar.murugesan@taylors.edu.my

## 1. INTRODUCTION

Edge computing is a distributed computing paradigm that brings computation and storage closer to the proximity of edge devices. These benefits have fueled many use cases like Artificial Intelligence (AI), robotics, machine learning and telco network communications and solved key challenges like bandwidth, latency, resilience, and data sovereignty. The motive of edge computing is to provide a decentralized cloud with low latency computation, overcome resource limitations of edge devices, and deal with network traffic and data explosion[1]. Figure 1 shows an edge computing framework that includes edge devices, edge servers and gateways as the essential components of the edge computing layer. Any device with computing, network and storage capability can act as an edge device. They gather data from the Internet of Things (IoT) devices, perform real-time data analysis, and respond faster than cloud computing. The gateways are responsible for translation service between various heterogeneous devices in the edge and IoT and cloud layers. The edge servers manage several edge devices and gateway by handling the context data. It is a generic term that captures associated computing paradigms such as fog computing, cloudlet, or mobile access edge computing[2].

Although edge computing provides many benefits, real-time data analytics that relies on edge computing faces various security and privacy issues. Edge data analytics includes data collection from different IoT devices, storing, processing, and analyzing them on geographically distributed edge nodes. The data owners lose control over the data transmitted to the edge nodes, and even achieving centralized control over them can be challenging. During this, the edge nodes may become vulnerable to intruders who can control the network, compromise the edge nodes, alter, or modify the transmitted data. These actions of the intruders can create integrity issues that result significantly in degrading the efficiency of edge data analytics and the performance of edge-based applications [3]. Therefore, an efficient data security solution in edge computing is required to ensure data integrity during edge data analysis and retain the performance of edge-based applications. Currently, there are many data integrity verification solutions to secure the data in edge computing, such as using blockchain devices, data assessments, collaborative methods or statistical anomaly detections[4]. These solutions increase resources utilization of lightweight edge nodes, adds computational load, or develop frequent interaction with intruders to understand their behaviours. Hence, the proposed research intends to address these issues and develop a lightweight security solution that does not drain the resource-limited edge nodes.

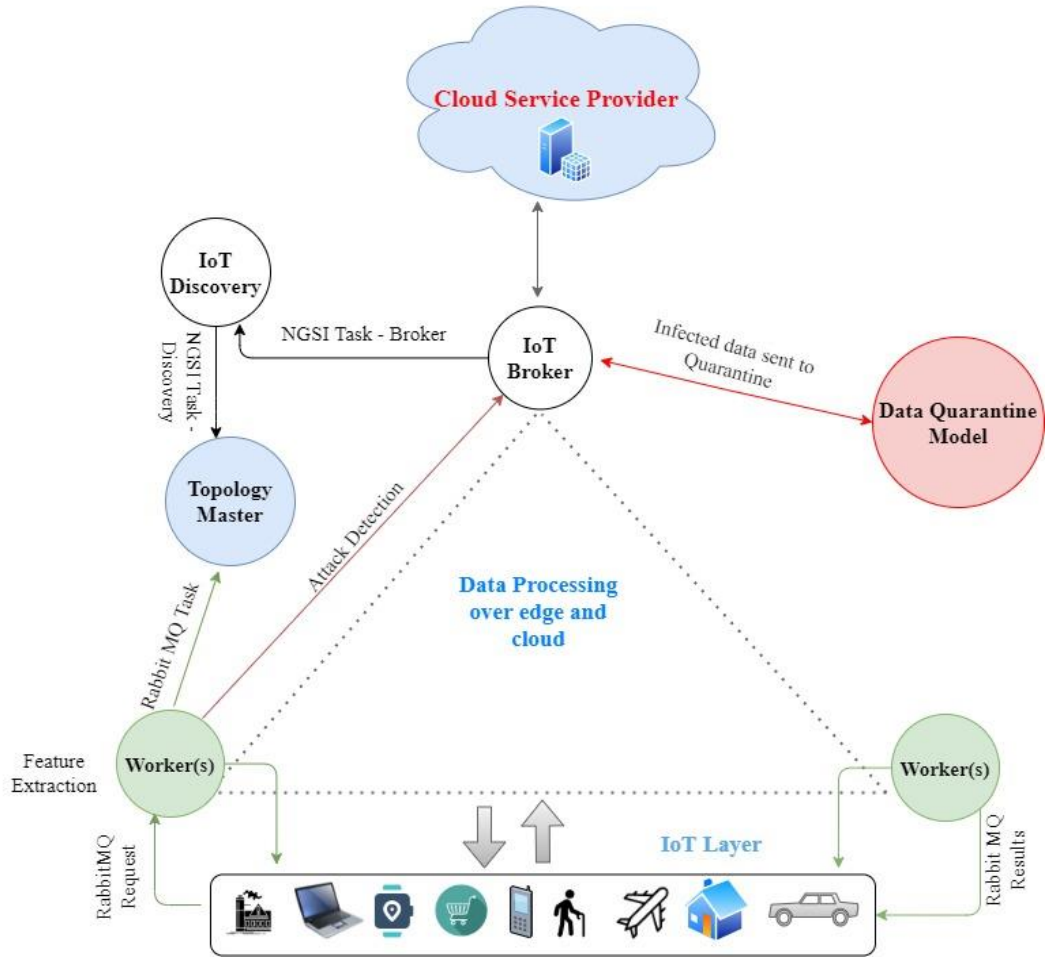

Figure 1. Edge computing framework with the dataflow

The proposed paper addresses data integrity issues in edge computing by identifying any intrusions caused in the network and later quarantine all the suspected devices and their data. The combined intrusion detection and data quarantine model identify the legitimate devices through the reputation score. Further, the confirmed legitimated devices data is sanitized based on the scrubbing score to recover the original data affected due to intrusion and thereby regain data integrity. The cyber security systems have used the same concept to quarantine the infected nodes from the computer system till their recovery[5]. In the cloud computing system, real-time data security using quarantine methods for petabytes of data from viruses and trojans is successful. In addition, there are many patent models for data security in file systems, databases, and computer systems. Due to these reasons, the proposed work adopts a data quarantine model to address the data security issues in edge computing. It is important to note that the quarantine approach is a renewed concept in edge computing. The proposed method does not add any computational load to the resource-constrained edge nodes, create alarms or allow suspected devices to interact with the other edge resources.

In the proposed concept, the quarantined devices will not be aware of the isolation and transmit data continuously. The spam detection module here uses this transmitted data to analyze their behaviour and identify legitimate devices. The data scrubber module sanitizes the data of these legitimate devices and sends it to the edge nodes. The sanitized data recovers the original data that was affected due to intrusion and assist in regaining the data integrity. As a preliminary work for the proposed concept, this paper identifies an appropriate Machine Learning (ML) for intrusion detection and quarantine. Overall, the objectives of the proposed concept are identified as follows:

- To identify the attack and isolate the infected data efficiently without alarming the nearby edge nodes.
- To prevent the illegitimate users from sending data to the edge nodes.
- To cleanse the inaccurate data and formulate sanitized data for edge data analytics.

The rest of the paper is organized as follows. Section 2 provides an overview of edge data analytics and data security issues in edge computing. Section 3 discusses the existing literature review. Section 4 presents the research methodology of the proposed concept. Section 5 evaluates the preliminary results. Section 6 discusses the obtained results, and finally, a conclusion with future work is included.

### 1.1. Motivation

The decentralized edge computing paradigm complements cloud computing by optimizing the performance of user-driven and communication-intensive applications. These features have fascinated many IoT applications to adopt edge computing and benefit from it. Smart Fog Hub Services is an edge-to-cloud platform deployed in Cagliari Airport in 2019 to provide a user-friendly and indoor map to the passengers to get their way to preferred shops, restaurants or any services during their waiting time in the airport [6]. This application promises the best customer experience and increases revenue for the airport and service providers. They use a recommender engine to track the users to infer their tastes and preferences based on their actions or similarities [7]. Many airports like Singapore provide smart airport services like biometric passports, Radio Frequency Identification (RFID) to track baggage, use beacons to track passengers, and many more. There are around 28% of smart airports in the world to augment users experience [8]. But this experience may create new cybersecurity challenges. There are many incidences wherein September 2018, British Airways was fined £183Million for a data breach in a security system that affected 380,000 transactions [9]. In May 2018, a data hack in Iran's Mashhad airport displayed protest messages in airport monitor's [10]. The hacking was mainly due to a privilege escalation attack. Therefore, when adopting these technologies and storing or sharing the data at the user's proximity, it is necessary to adopt security measures to safeguard data confidentiality and integrity. For instance, these data hacks by spammers or attackers can introduce undesirable data and strain into the system.

### 2. EDGE COMPUTING AND DATA SECURITY ISSUES IN EDGE LAYER

The standard computing framework, FogFlow developed by the NEC laboratory, is used as a ground framework, and deploy the proposed concept. This framework supports standard interfaces to share and reuse contextual data across services. As shown in Figure 1, there are three logical divisions in the framework: service management, data processing, and context management. The service management includes Topology Master (TM), task designer and docker image repository. Task designer provides a web interface to monitor the IoT services, and the docker repository manages all the docker images. TM is responsible for service orchestration to handle service requirements and service topology among the edge nodes. The worker or edge nodes at the proximity of IoT devices perform data processing tasks assigned by TM. TM and workers communicate through RabbitMQ protocol. Finally, context management includes IoT discovery, a set of IoT brokers and federated brokers. These components establish data flow across the tasks and manage contextual data like availability of workers, topology, task and generated data stream [11]. Figure 1 also depicts the process and data flow of the framework. The FogFlow framework supports various use cases like lost child finder, anomaly detection in smart cities, smart parking, and smart industry. Therefore, implementing the proposed Data Quarantine Model (DQM) considering the FogFlow framework can support the real-time applicability of the concept to address data integrity issues in any use case scenarios.

### 2.1 DATA SECURITY ISSUES

Edge computing handles data from various sensors, devices, servers, including local data centres and centralized cloud. The edge nodes gather data from the different ubiquitous devices, transmit it to the other edge nodes or servers, and analyze it. The data analysis includes task deployment defined as a mobile agent and deployed dynamically on the edge nodes. These tasks can be parallel, asynchronous and sometimes independent of one another, without the intervention of the other nodes [12]. The data collection, transmission, and task deployment process provide appropriate real-time interaction and monitoring of the applications, known as edge data analytics. However, decentralized edge data analytics are vulnerable to security threats due to a lack of centralized control and the features of edge nodes such as heterogeneousness, mobility, and geographical distribution. The vulnerabilities can create various attacks that affect data confidentiality, integrity, authentication, and access control. Figure 2 shows various threats that cause these data security issues.

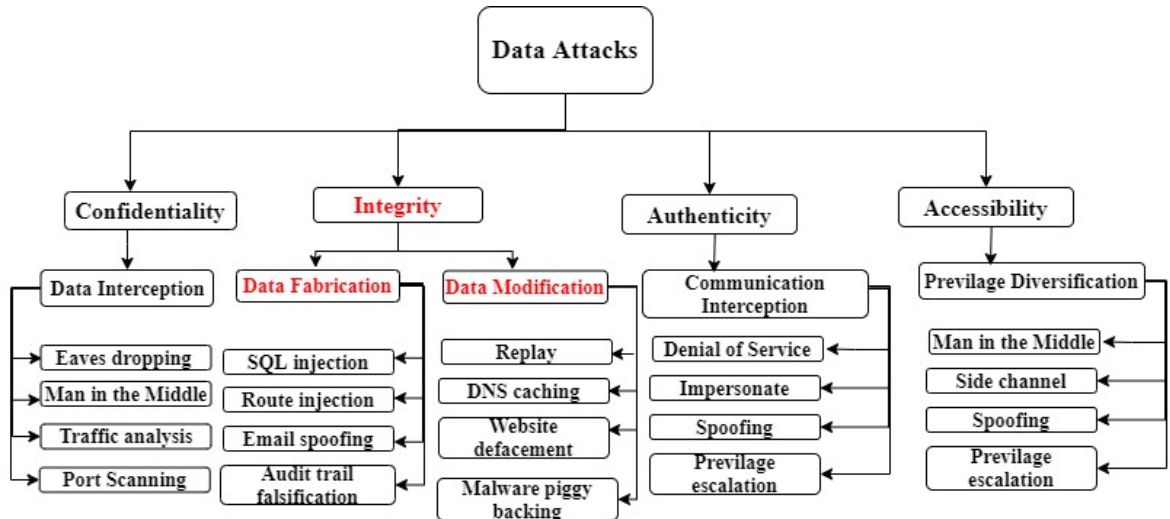

Figure 2. Category of Data Attack in Edge Computing

**Data Confidentiality** – It is a fundamental requirement in the edge layer to ensure that the data owners and users can access private information without the intervention of unauthorized users. The edge nodes receive data in the core network infrastructure, transmit and process it to the other edge nodes or cloud data centres [13]. The attackers can capture the data packets during communication and read sensitive information like passwords, usernames, and credit card details during this process. If one or more edge nodes are affected, the adversaries can wiretap the entire system and gradually decode the data packets [14]. Therefore, the security frameworks must manage and store data without compromising the confidentiality in the edge layer. Common approaches used to achieve data confidentiality are cryptography schemes, secure data acquisition and secure IoT devices or edge nodes. Dynamic keys encrypt, authenticate, and fragment input data in cryptography schemes. In a secure data acquisition, retransmission of decisions for each communication optimizes the data. Apart from this, Blockchain or Artificial Intelligence (AI) are incorporated on IoT devices or edge nodes to validate the data periodically.

**Data Integrity** - This is a measure to refer originality, completeness, and accuracy of the data from the sourcing until the entire data analysis process. The data owners lose control over their data while outsourcing to the edge nodes, and thereby the data can be vulnerable to any security attacks. The attacks can modify, alter, or delete the data with malicious intent creating integrity issues. [15]. Lack of data integrity and correctness solution can affect the performance and efficiency of edge computing. The possible solutions to achieve data integrity are batch auditing or dynamic auditing using homomorphic tags to the outsourced data. There are few privacy-preserving protocols to enhance security through monitoring the system.

**Data Authentication** – It is the process to validate the identity of users and guarantee that the users are legitimate to access cloud servers or edge nodes. In addition, it is necessary to authenticate the participating edge nodes and edge data centres. By establishing authentication in edge computing, data quality can be improved drastically [16]. Password, smart card, biometrics-based authentication are the typical authentication schemes adopted in cloud and edge computing applications.

**Accessibility**– This is a mechanism for determining the rights and privileges of edge nodes and users in the system. All the users in the network cannot access sensitive information. Hence, it is necessary to specify local policies to determine access specifications for users and resources in the infrastructure. In edge computing, choosing accessibility can be for three reasons: storage and computation services, edge nodes to access particular resources, and virtual machines [17]. Some of the access control mechanisms adopted in edge computing are access control models like – Mandatory access control, discretionary access control, Role-based access control and Attribute-based access control.

Among all the above data attacks, the data integrity issues severely impact the data stored and processed in edge computing. The issues can lead to inconsistent and unreliable information leading to the wrong decision during edge data analytics. Therefore, the proposed method ensures integrity issues caused due to any intrusion resulting in data modification, alteration, or injection to the transmitted data in edge computing and present a concept to handle these issues. False data injection, SQL or route injection, spam, data falsification, malware, replay are some threats that cause data integrity issues through data fabrication or modification [18]. It is necessary to address these issues immediately since their impact can propagate from one edge node to another and reduce the performance by affecting the data analysis process. In addition, it would be challenging to identify the root cause and results in additional repair costs and delay recovery[19]. Hence, we propose an analytical method to detect integrity attacks due to false data injection by employing a data quarantine model.

## 3. LITERATURE REVIEW

Data integrity attack or integrity attack is an attempt to corrupt the data intentionally by modifying, deleting, or fabricating during data outsourcing. Data, route or SQL injection, malware, email spoofing, replay are the possible attack vectors to affect data integrity. Through this attack, the attackers can understand the communication protocol and gain control over the system resulting in unstable operations or prolonged loss of resources. The impact of the attack is proportional to the duration of the attack[20]. There are limited works in edge computing to address data integrity issues, and Table 1, summarizes these related works. Blockchain is a decentralized ledger that manages multiple participants collaboratively without centralized control. Both edge computing and blockchain have distributed architecture; hence adopting it here can help achieve significant interdependency. In distributed neural networks, blockchain tracks end-to-end IoT based applications. An efficient inter-blockchain querying and locking mechanism in the edge layer improves security and performance [21].

Similarly, blockchain is adopted to ensure data integrity during the task offloading process. They employ simple additive weighing and multi-criteria decision-making techniques to confirm the optimal migration of virtual machines. But achieving optimal migration requires many iterations, which is time-consuming [22]. Despite the advantage of blockchain providing centralized monitoring of the application, they are resource hungry and immutable for stored data. These features of blockchain can be a great threat to edge computing applications; however, further research in this area can solve many issues.

Table 1. Related work

| Reference | Working model | Attacks addressed | Limitation |
|---|---|---|---|
| [21] | Blockchain-enabled integrity protection scheme in distributed neural networks | Data poisoning and model poisoning | Blockchain increases resource utilization of lightweight edge nodes |
| [22] | Blockchain-enabled integrity preserving system during task offloading | Unauthorized transaction | Blockchain increases resource utilization and makes stored data immutable |
| [23] | Collaborative bad data detection scheme in the electric metering system | False data injection destroying data integrity and availability | Increases computational load to edge nodes |
| [24] | Damage assessment and data recovery to ensure consistent database in smart city applications | Malicious activity and privacy violation | Increases computational load to edge nodes |
| [25] | A statistical learning-based anomaly detection system in smart grid application | Data integrity attack due to false data injection | Additional computational load to track disseminated data is included |

A collaborative bad data detection scheme identifies integrity issues due to false data injection in an electric metering system. It includes a set of rules on the edge nodes to determine the reputed electric meter. These rules identify false data injection that destroys the integrity and availability of the data [23]. Damage assessment and data recovery system for smart city applications is another solution to recover data from malicious attacks. This solution recovers original data and returns the database to a consistent state for data analysis in edge computing. This approach has three recovery algorithms: main damage assessment, secondary damage assessment, and main recovery for each data transaction. This process continues till all the affected transactions in the entire system are detected [24]. This approach includes detection schemes on each edge nodes and increase the computational time of lightweight edge nodes.

Anomaly detection for data integrity attack in smart city application is used to secure data and privacy in the edge layer. Edge nodes in the higher hierarchical layer that has access to inter-area data have anomaly detectors. They regularly scan the input data stream to ensure data integrity in the entire system [25]. Although this approach provides a conventional centralized approach to obtain results with reliable decisions, guaranteeing the data integrity can be a tedious process. Edge nodes at the lower level would have disseminated the data to other edge nodes, so tracking the data will add computation to the edge nodes. Overall, the discussed security solutions have not considered mitigating the attack early during data acquisition. Addressing the issue during data acquisition can assist in data tracking and identify the root cause of the issue.

## 4. RESEARCH METHODOLOGY

This paper proposes a data quarantine model to defend against data integrity attacks. The proposed model considers continuous measures to quarantine the affected data without affecting data analysis in the edge layer. Figure 1 shows intrusion detection includes in the existing FogFlow framework through feature extraction. The worker nodes fetch the data from the IoT devices, analyze data and identify any intrusions. After identifying the intrusions on the edge nodes, the edge server - IoT broker quarantines the infected data for a predetermined time in DQM. Figure 3 shows the modules in the DQM, and the following section includes the description of each module.

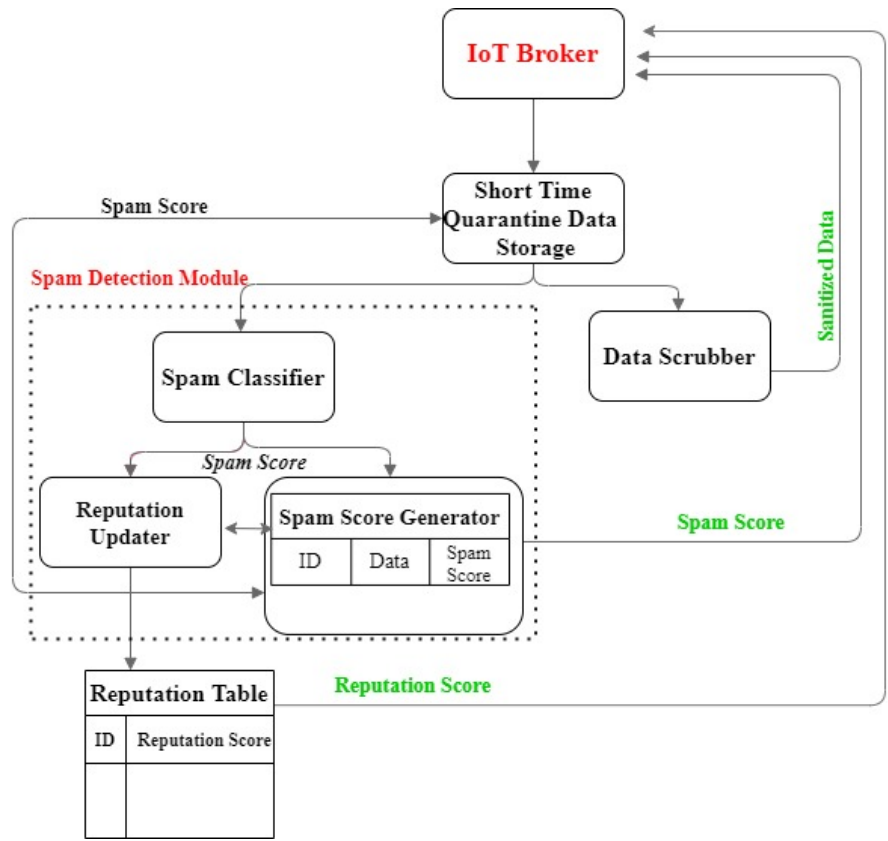

Figure 3. Proposed Data Quarantine Model

### 4.1 Short Time Quarantine Data Storage

Short time quarantine data storage is a temporary database to store intruded devices and their data. The storage includes device ID, device location, worker ID, and the sourced data. This data is passed to the spam detection module to identify spam and generate the devices reputation and spam scores. This data storage stores these scores that assist in predicting the behaviour of the devices. Later passes it to the data scrubbing module to perform data sanitization. This temporary data storage stores all these scores for a predetermined time until the edge resources receive the sanitized data. The untrusted data and devices information is discarded to retore the data storage.

### 4.2 Spam Detection Module

The spam detection module analyses the behaviour of the devices based on the data they transmit during quarantine. The spam classification matrix categorizes the data as spam or non-spam using and Table 2 shows the spam confusion matrix. The confusion matrix includes the following values: TP – (True Positive) represents the spam correctly classified, FN – (False Negative) spam misclassified as a spammer, FP – (False

Positive) Non-spammer misclassified, and TN – (True Negative) is the number of non-spammers classified correctly. The spam score generator generates the spam score for the classified spam data using equation (1).

Table 2. Confusion Matrix

|  |  | True class | |
|---|---|---|---|
|  |  | Positive | Negative |
| Predicted class | Positive | TP | FP |
|  | Negative | FN | TN |

To obtain spam scores the below equation is used

$$SS = \frac{(NSD + (NMD \times NSS))}{SI} \quad (1)$$

Where, SS – **S**pam **S**core, NSD – **N**umber of attacks sent to the **S**ame **D**estination, NMD – **N**umber of attacks sent to **M**ultiple **D**estination, NSS – **N**umber of attacks **S**ent by **S**ingle attacker, and SI – **S**pam **I**nterval

In equation (1), SI determines the time interval between the two attacks. If the time interval between two attacks is small, the spam score is higher. If the attackers send the attack to the same or different destination, then NSD and NMD increase. Similarly, if a single attacker sending attacks to many destinations, NSS is considered. Based on this score, a SS of 9 is considered a threshold and used in reputation updater to identify legitimate devices.

**4.3 Reputation Updater**

Reputation updater and spam score is bi-directional feedback to each other. It includes a reputation table with device id and reputation score. Based on the spam score, the reputation of each device is measured. Figure 4 shows the relation between these scores in the spam detection module. If the devices have a spam score greater than the threshold, they are blacklisted devices, and IoT brokers can prevent receiving data from blocked devices. If the past data in the spam module has the least spam score, then the devices are considered a whitelist and are authorized to transmit data to the edge layer.

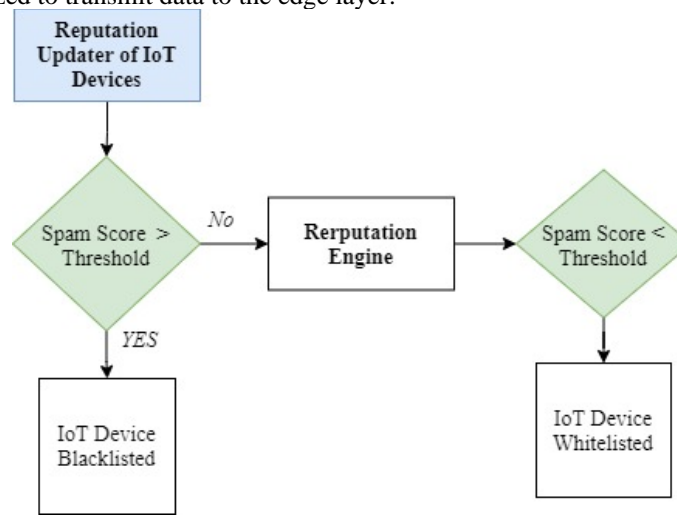

Figure 4. Illustration of Reputation Score

**4.4 Data Scrubber**

The data scrubber module sanitizes the inconsistent, uncertain, and ambiguous trusted data of legitimated devices stored in a short time data storage. First, they prefetch the data from the spam classifier to convert and consolidate it into a single format. Later several rules are applied to validate the consistency of the data using scrubbing scores. Finally, the data is loaded back to the database and sent to the IoT broker for edge data analytics. The sanitized data obtained improves the data quality, integrity and enhance the edge decision support system.

## 5  RESULTS

The proposed work implemented intrusion detection on the worker edge nodes during data acquisition as preliminary work. The worker nodes process the acquired data to obtain a high-degree spatial separation of high dimensional data. This dimensional reduction monitors the input data stream and updates any changes to the IoT broker. The main objective of updating changes is to identify any attacks and quarantine the suspected devices and their data. IoT broker decides to pass the data either to the quarantine model or the available edge nodes for data analysis based on the signs of a possible attack. This whole approach of data quarantine and indicating attacks in the devices do not add any load to the existing edge nodes or create notifications to other resources.

### 5.1  Dimensionality Reduction

Dimensionality reduction is a Machine Learning (ML) technique to reduce the higher dimensional data to lower dimensions that remove redundant and dependent features. To identify the efficient ML method for feature reduction, the following ML methods are analyzed, like Linear Discriminant Analysis (LDA), Logistic Regression (LR), and Support Vector Machine (SVM) and nonlinear classifier Multi-Layer Perceptron (MLP). The comparative study considers the NSL-KDD dataset and evaluates these methods in terms of training time, training accuracy and quarantine accuracy

### 5.2  Experimental Setup

The proposed work implemented the ML dimensionality reduction techniques on Python-based YAFS (Yet Another Fog Simulator) [26] using the NSL-KDD dataset[27]. The same tool will be used to implement the proposed model in future. The dataset consists of varying proportions of normal and attack data set. The simulation setup includes a cloud node of 10GB RAM and 16GHz CPU, two edge nodes of 2GB RAM and 3 GHz CPU, and 50-400 IoT devices of 500MB RAM and 1 GHz CPU power. The links bandwidth is 3-10 Mbits, the packet size of 200 bytes and $100 \, X \, 10^8$ packets per instructions.

Figure 5 shows the results obtained by the feature extractions using the methods mentioned above. Figure 5(a) shows that LDA and LR has a minimum training time of 0.9seconds and 1.5seconds, respectively since LDA is simple, efficient and has less computational overhead. The LDA class matrix of dimensionality reduction is inverse of the grouped covariance matrix, which transposes input data and class mean vectors that make it very simple. Similarly, in LR, a normalization technique is applied to achieve the objective using the sigmoid function. Figure 5(b) shows the training accuracy of these methods and notice that MLP has 99.64% accuracy, but the training time is 88.95seconds which is considerably more than LD and LR. MLP is a particular case of supervised Artificial Neural Networks (ANN). This network architecture includes an input layer, one or more hidden layers with some number of neurons, and an output layer with one neuron for each class to be classified. Therefore, it is possible to achieve higher accuracy by comparing each value of the distribution vectors

The proposed quarantine aims to send infected nodes to the quarantine, and hence LDA and LR techniques are further analyzed to achieve better precision towards devices and data quarantine. Quarantine is forced isolation or stoppage of the interaction of the infected device and data with the edge nodes in the framework. Figure 5 (c) shows the quarantine accuracy of LDA and LR techniques, and it shows that LDA has 72.83% accuracy in sending the devices to the quarantine section.

### 5.3 Evaluating the devices and data sent to Quarantine

As noted, LDA and LR has better quarantine accuracy compared to SVM and MLP. This section analyses LDA and LR to identify efficient quarantine techniques. The following equation determines the accuracy of these methods:

$$\frac{TP + TN}{TP + TN + FP + FN}$$

where,
TP (True Positive): Number of instances correctly classified as an attack.
FP (False Positive): Number of instances wrongly classified as an attack.
TN (True Negative): Number of instances correctly classified as normal
FN (False Negative): Number of instances wrongly classified as an attack.

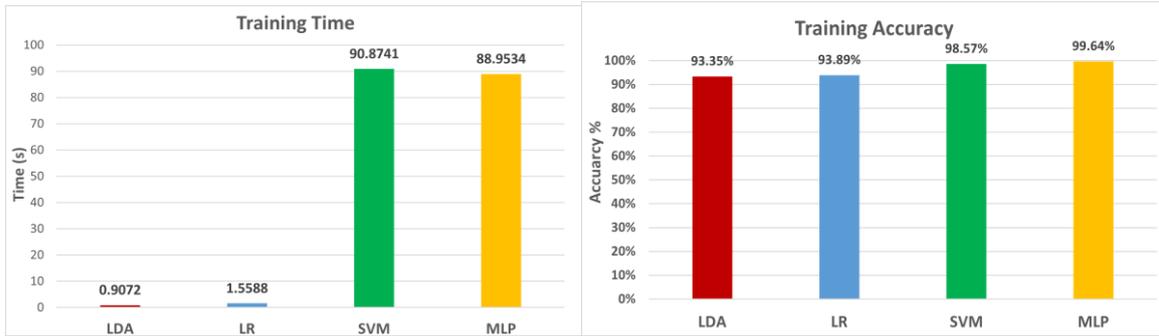

Figure 5 (a). Training Time        Figure 5 (b). Training Accuracy

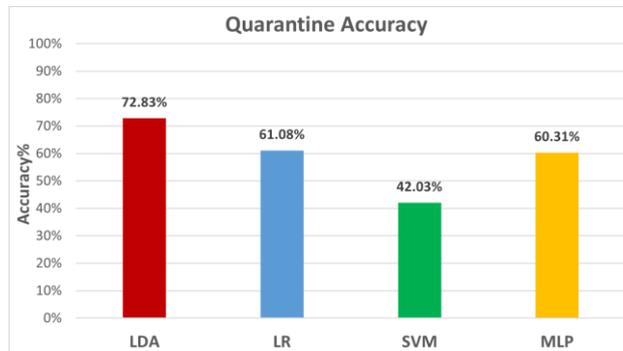

Figure 5 (c). Quarantine Accuracy

Figures 6(a) and 6(b) show that the LDA quarantines 265 devices out of 400 and 2047 data packets, respectively. Figure 6(c) shows that the quarantine accuracy of LR with 50 IoT devices is slightly lesser compared with the higher number of IoT devices. But the accuracy of LDA is 72% irrespective of the number of IoT devices. This quarantine accuracy shows that LDA is more efficient and reliable in quarantining the data and IoT devices regardless of the number of devices. Therefore, using LDA techniques, the proposed DQM will be implemented further and address the data integrity issues caused due to any intrusions in edge computing.

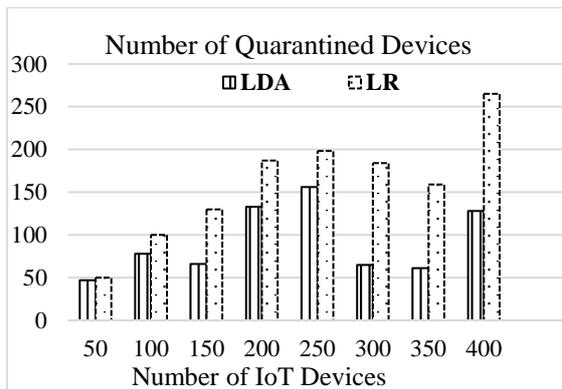
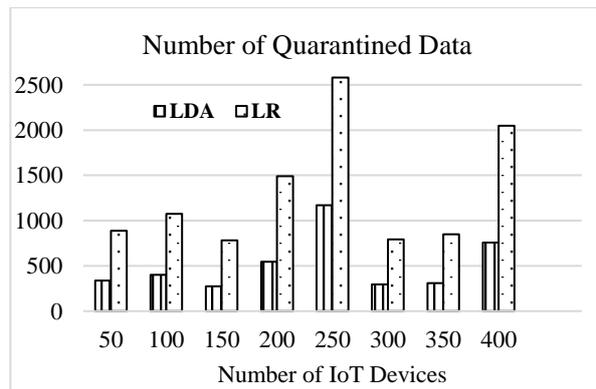

Figure 6 (a). Devices Quarantine        Figure 6 (b). Data Quarantine

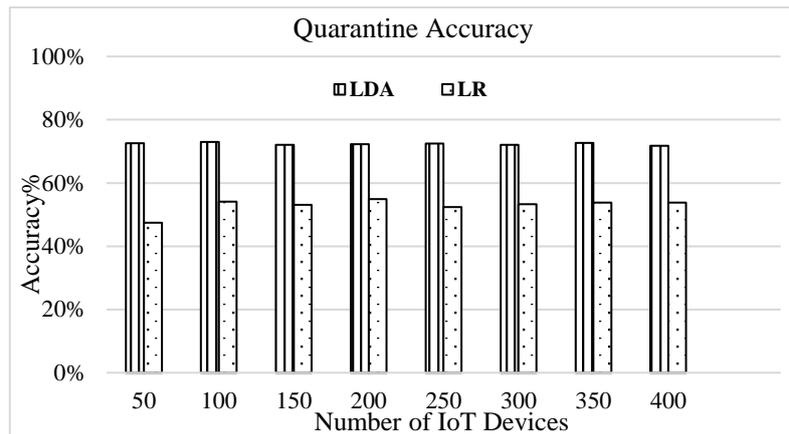

Figure 6 (c). Quarantine Accuracy

## 5. DISCUSSION

The preliminary results obtained in the proposed work includes the ML methods for intrusion detection. The worker nodes in the FogFlow framework perform the dimensionality reduction of the data to identify any intrusive activities. The training time of 0.9 s and 1.5 s of LR and LDA methods shows that these methods identify intrusions faster. The faster detection would limit the attacker presence in edge resources and do not provide any possibilities for the attackers to understand the behaviour of the framework. Subsequently, this can significantly reduce the intensity of the attack on the framework. Although the considered ML methods have more than 90% accuracy, the proposed research focuses on quarantining the affected devices. Hence, while analyzing the quarantine efficiency, LDA provides constant 72% accuracy regardless of the number of IoT devices. Therefore, LDA was considered to have a better quarantine accuracy and efficient training time of 0.9 seconds compared to other methods.

## 6. CONCLUSION AND FUTURE WORK

Edge computing is a promising architecture that provides a data processing platform between cloud and edge nodes. Communication intensive and latency-sensitive applications have more significant benefits due to edge computing. But security and privacy issues are challenging when the computation is brought closer to the edge nodes. It can be observed from the literature that data attacks can compromise the edge nodes to gain access into the network and affect data confidentiality, integrity, availability, and access control. Among these attacks, data integrity attacks can severely impact the efficiency and reliability of edge computing systems. These integrity issues are mainly because the integrity attack causes inconsistent and unreliable data that contributes to the distressing edge data analysis process. Therefore, it is crucial to address the data integrity attack early before it propagates to the other edge nodes. Failure of which can lead to challenging situations to track the root cause of the attack. All the existing frameworks adopt a security framework to handle data integrity attacks by including anomaly detections on the existing edge nodes or adding blockchain to identify the attack. These mechanisms can increase the computational load of edge nodes affecting the lightweight feature of edge nodes. The proposed work uses a lightweight dimensionality reduction technique to monitor intrusions and quarantine the data and IoT devices. In future work, this technique will be improved to achieve better quarantine accuracy.

This research proposes a new concept of data quarantine model to guarantees data integrity in edge computing. This model can adaptively identify the degree of intervention on the edge data while sourcing the data from IoT devices. The identified data and devices are alleviated to quarantine model for a predetermined time without alarming the data analysis process of the neighbouring edge nodes or by adding computational load. The spam detection module identifies the legitimate devices, and their data is passed to the data scrubber module. In the scrubber module, sanitized data is passed to the edge layer to continue data analysis. The data quarantine model shall be implemented and validated with ground truth data for efficiency estimation in future work.

**ACKNOWLEDGEMENT**

This research work is supported by Taylor's University, Malaysia, through its Taylor's PhD Scholarship Program.

**REFERENCES**

[1] A. M. Alsmadi *et al.*, "Fog computing scheduling algorithm for smart city," *Int. J. Electr. Comput. Eng.*, vol. 11, no. 3, pp. 2219–2228, 2021, doi: 10.11591/ijece.v11i3.pp2219-2228.

[2] M. El Ghmary, Y. Hmimz, T. Chanyour, and M. O. Cherkaoui Malki, "Time and resource constrained offloading with multi-task in a mobile edge computing node," *Int. J. Electr. Comput. Eng.*, vol. 10, no. 4, pp. 3757–3766, 2020, doi: 10.11591/ijece.v10i4.pp3757-3766.

[3] M. Mukherjee *et al.*, "Security and Privacy in Fog Computing: Challenges," *IEEE Access*, vol. 5, pp. 19293–19304, 2017, doi: 10.1109/ACCESS.2017.2749422.

[4] S. Khan, S. Parkinson, and Y. Qin, "Fog computing security: a review of current applications and security solutions," *J. Cloud Comput.*, vol. 6, no. 1, 2017, doi: 10.1186/s13677-017-0090-3.

[5] V. Coskun, E. Cayirci, A. Levi, and S. Sancak, "Quarantine region scheme to mitigate spam attacks in wireless sensor networks," *IEEE Trans. Mob. Comput.*, vol. 5, no. 8, pp. 1074–1086, 2006, doi: 10.1109/TMC.2006.121.

[6] A. Salis, G. Mancini, R. Bulla, P. Cocco, D. Lezzi, and F. Lordan, "Benefits of a Fog-to-Cloud Approach in Proximity Marketing," in *Lecture Notes in Computer Science (including subseries Lecture Notes in Artificial Intelligence and Lecture Notes in Bioinformatics)*, vol. 11339 LNCS, 2019, pp. 239–250.

[7] M. Aljarah, M. Shurman, and S. H. Alnabelsi, "Cooperative hierarchical based edge-computing approach for resources allocation of distributed mobile and IoT applications," *Int. J. Electr. Comput. Eng.*, vol. 10, no. 1, pp. 296–307, 2020, doi: 10.11591/ijece.v10i1.pp296-307.

[8] G. Lykou, A. Anagnostopoulou, and D. Gritzalis, "Smart Airport Cybersecurity: Threat Mitigation and Cyber Resilience Controls," *Sensors*, vol. 19, no. 1, p. 19, Dec. 2018, doi: 10.3390/s19010019.

[9] "British Airways hack_ Credit card details of 380,000 stolen." 2018.

[10] Times of Israle, "Screens at Iran airport said hacked with anti-regime messages _ The Times of Israel." 2018.

[11] B. Cheng, G. Solmaz, F. Cirillo, E. Kovacs, K. Terasawa, and A. Kitazawa, "FogFlow: Easy Programming of IoT Services Over Cloud and Edges for Smart Cities," *IEEE Internet Things J.*, vol. 5, no. 2, pp. 696–707, Apr. 2018, doi: 10.1109/JIOT.2017.2747214.

[12] C. L. Stergiou, K. E. Psannis, and B. B. Gupta, "Iot-based big data secure management in the fog over a 6G wireless network," *IEEE Internet Things J.*, vol. 8, no. 7, pp. 5164–5171, 2021, doi: 10.1109/JIOT.2020.3033131.

[13] J. Zhang, B. Chen, Y. Zhao, X. Cheng, and F. Hu, "Data Security and Privacy-Preserving in Edge Computing Paradigm: Survey and Open Issues," *IEEE Access*, vol. 6, no. Idc, pp. 18209–18237, 2018, doi: 10.1109/ACCESS.2018.2820162.

[14] P. Mahadevappa and R. K. Murugesan, "Study of Container-Based Virtualisation and Threats in Fog Computing," vol. 1347, M. Anbar, N. Abdullah, and S. Manickam, Eds. Singapore: Springer Singapore, 2021, pp. 535–549.

[15] R. G. Engoulou, M. Bellaïche, S. Pierre, and A. Quintero, "VANET security surveys," *Comput. Commun.*, vol. 44, pp. 1–13, May 2014, doi: 10.1016/j.comcom.2014.02.020.

[16] D. Liu, Z. Yan, W. Ding, and M. Atiquzzaman, "A Survey on Secure Data Analytics in Edge Computing," *IEEE Internet Things J.*, vol. 6, no. 3, pp. 4946–4967, Jun. 2019, doi: 10.1109/JIOT.2019.2897619.

[17] P. Zhang, J. K. Liu, F. Richard Yu, M. Sookhak, M. H. Au, and X. Luo, "A Survey on Access Control in Fog Computing," *IEEE Commun. Mag.*, vol. 56, no. 2, pp. 144–149, 2018, doi: 10.1109/MCOM.2018.1700333.

[18] A. Zainab, S. S. Refaat, and O. Bouhali, "Ensemble-Based Spam Detection in Smart Home IoT Devices Time Series Data Using Machine Learning Techniques," *Information*, vol. 11, no. 7, p. 344, Jul. 2020, doi: 10.3390/info11070344.

[19] M. El Ghmary, T. Chanyour, Y. Hmimz, and M. O. C. Malki, "Efficient multi-task offloading with energy and computational resources optimization in a mobile edge computing node," *Int. J. Electr. Comput. Eng.*, vol. 9, no. 6, pp. 4908–4919, 2019, doi: 10.11591/ijece.v9i6.pp4908-4919.

[20] S. Tuuli and R. Mika, "Detecting Stuxnet-like data integrity attacks," *Secur. Priv.*, vol. 3, no. 5, pp. 1–10, 2020, doi: 10.1002/spy2.107.

[21] G. S. Aujla *et al.*, "BloCkEd: Blockchain-based secure data processing framework in edge envisioned V2X environment," *IEEE Trans. Veh. Technol.*, vol. 69, no. 6, pp. 5850–5863, 2020, doi: 10.1109/TVT.2020.2972278.

[22] X. Xu, X. Zhang, H. Gao, Y. Xue, L. Qi, and W. Dou, "BeCome: Blockchain-Enabled Computation Offloading for IoT in Mobile Edge Computing," *IEEE Trans. Ind. Informatics*, vol. 16, no. 6, pp.

4187–4195, 2020, doi: 10.1109/TII.2019.2936869.
[23]  Z. Cai, B. Qian, and Y. Xiao, "Edge Computing Based Bad Metering Data Detection," *2019 3rd IEEE Conf. Energy Internet Energy Syst. Integr. Ubiquitous Energy Netw. Connect. Everything, EI2 2019*, pp. 693–698, 2019, doi: 10.1109/EI247390.2019.9062052.
[24]  A. Alazeb and B. Panda, "Maintaining data integrity in fog computing based critical infrastructure systems," *Proc. - 6th Annu. Conf. Comput. Sci. Comput. Intell. CSCI 2019*, pp. 40–47, 2019, doi: 10.1109/CSCI49370.2019.00014.
[25]  M. Davoodi, R. Moslemi, W. Song, and J. M. Velni, "A fog-based approach to secure smart grids against data integrity attacks," *2020 IEEE Power Energy Soc. Innov. Smart Grid Technol. Conf. ISGT 2020*, 2020, doi: 10.1109/ISGT45199.2020.9087790.
[26]  I. Lera, C. Guerrero, and C. Juiz, "YAFS: A Simulator for IoT Scenarios in Fog Computing," *IEEE Access*, vol. 7, pp. 91745–91758, 2019, doi: 10.1109/ACCESS.2019.2927895.
[27]  C. I. for Cybersecurity, "NSL-KDD | Datasets | Research | Canadian Institute for Cybersecurity | UNB." https://www.unb.ca/cic/datasets/nsl.html (accessed Apr. 20, 2020).

**BIOGRAPHIES OF AUTHORS**

| | |
|---|---|
| 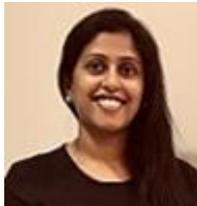 | Poornima Mahadevappa is currently a PhD Scholar in Computer Science at Taylor's University, Malaysia. She obtained her Master of Engineering in Bioinformatics (2014) from UVCE, Bangalore University, India, and Bachelor of Engineering in Computer Science (2008) from Ghousia College of Engineering, Visvesvaraya Technological University, India. She worked as a lecturer in Pooja Bhagavat Memorial Mahajana Post Graduate Centre and as an Android developer previously. Her research interest includes edge computing, cyber security, and data analytics.<br>Email: poornimamahadevappa@sd.taylors.edu.my<br>ORCID: 0000-0001-9414-3464;<br>Google Scholar: Poornima Mahadevappa |
| 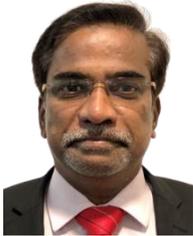 | Dr Raja Kumar Murugesan is an Associate Professor of Computer Science, and Head of Research for the Faculty of Innovation and Technology at Taylor's University, Malaysia. He has a PhD in Advanced Computer Networks from the Universiti Sains Malaysia. His research interests include IPv6, Future Internet, Internet Governance, Computer Networks, Network Security, IoT, Blockchain, and Machine Learning. He is a member of the IEEE and IEEE Communications Society, Internet Society (ISOC), and associated with the IPv6 Forum, Asia Pacific Advanced Network Group (APAN), Internet2, and Malaysia Network Operator Group (MyNOG) member's community. He has held various leadership roles in his academic career. Raja has given several invited talks and presentations on IPv6, Internet Governance, IoT, Blockchain, AI, Machine learning and Digital Transformation at various international conferences and events.<br>Email: rajakumar.murugesan@taylors.edu.my<br>ORCID: 0000-0001-9500-1361<br>Google Scholar: Raja kumar Murugesan |